\relax
\documentclass[11pt,a4paper]{article} 
\usepackage{authblk}
\usepackage[hyperref]{emnlp2018}  
\usepackage{times}  
\usepackage{latexsym}
\usepackage{url}  
\usepackage{graphicx}  
\usepackage{tabularx}
\usepackage{multirow}
\usepackage{booktabs}
\usepackage[normalem]{ulem}
\usepackage[utf8]{inputenc}
\usepackage{algorithm}
\usepackage{algpseudocode}
\usepackage{array}
\usepackage{amsmath}
\usepackage[justification=centering]{caption}

\usepackage{amsmath}
\usepackage{gb4e}

\noautomath 

\newcolumntype{L}[1]{>{\raggedright\let\newline\\\arraybackslash\hspace{0pt}}m{#1}}
\newcolumntype{C}[1]{>{\centering\let\newline\\\arraybackslash\hspace{0pt}}m{#1}}
\newcolumntype{R}[1]{>{\raggedleft\let\newline\\\arraybackslash\hspace{0pt}}m{#1}}

\useunder{\uline}{\ul}{}
\newcolumntype{Y}{>{\centering\arraybackslash}X}

\aclfinalcopy 
\begin{document}

\title{A Knowledge Hunting Framework for Common Sense Reasoning}

\author[1]{Ali Emami}
\author[1]{Noelia De La Cruz}
\author[2]{Adam Trischler}
\author[2]{Kaheer Suleman}
\author[1]{Jackie Chi Kit Cheung}
\affil[1]{School of Computer Science, Mila/McGill University}
\affil[2]{Microsoft Research Montreal}
\affil[ ]{\textit {\{ali.emami, noelia.delacruz\}@mail.mcgill.ca}}
\affil[ ]{\textit {\{adam.trischler, kasulema\}@microsoft.com}}
\affil[ ]{\textit {jcheung@cs.mcgill.ca}}

\maketitle

\begin{abstract}
We introduce an automatic system that achieves state-of-the-art results on the Winograd Schema Challenge (WSC), a common sense reasoning task that requires diverse, complex forms of inference and knowledge.
Our method uses a knowledge hunting module to gather text from the web, which serves as evidence for candidate problem resolutions.
Given an input problem, our system generates relevant queries to send to a search engine, then extracts and classifies knowledge from the returned results and weighs them to make a resolution.
Our approach improves F1 performance on the full WSC by 0.21 over the previous best and represents the first system to exceed 0.5 F1. We further demonstrate that the approach is competitive on the Choice of Plausible Alternatives (COPA) task, which suggests that it is generally applicable.

\end{abstract}

\section{Introduction}

The importance of common-sense reasoning in natural language processing, particularly for syntactic and semantic disambiguation, has long been recognized.
Almost 30 years ago, \citet{dahlgren1989knowledge} proposed systems that use common sense to disambiguate parse trees, word senses, and quantifier scope.
Although the resolution of certain ambiguities depends chiefly on linguistic patterns (e.g., the number and gender of an antecedent for pronoun disambiguation), many cases depend on world knowledge, shared points of reference, and an understanding of what is plausible---concepts often grouped under the term ``common sense.'' 

Various tasks have been devised to test common-sense reasoning in automatic systems.
Two of the most popular are the \emph{Winograd Schema Challenge} (WSC)~\cite{levesque2011winograd}~
and
the \emph{Choice of Plausible Alternatives} (COPA) \cite{roemmele2011choice}.
Both require a system to assess the relative plausibility of two scenarios.

WSC problems are short passages containing a target pronoun that must be correctly resolved to one of two possible antecedents. They come in pairs which differ slightly and result in adverse correct resolutions. As an example:
\begin{exe}
\ex \begin{xlist} 
    \ex \label{ex-1a} Jim \underline{yelled at} Kevin because \textit{he} was so upset. (Answer: Jim)
    \ex Jim \underline{comforted} Kevin because \textit{he} was so upset. (Answer: Kevin)
    \end{xlist}
\end{exe}
WSC problem pairs (``twins,'' using the terminology of \citet{hirst1988semantic}) are carefully controlled such that heuristics involving syntactic salience, the number and gender of the antecedent, or other simple syntactic and semantic cues are ineffective. This distinguishes the task from the standard coreference resolution problem. Performant systems must make common sense inferences; i.e., that someone who yells is likely to be upset, and that someone who is upset tends to be comforted. Additional examples are shown in Table~\ref{tab:examples}.
\begin{table*}[ht]
\small 
\begin{center}
\begin{tabular}{cL{13.5cm}} 
\hline

1 a) & The man couldn't lift his son because he was so \underline{weak}. (Answer: the man) \\
1 b) & The man couldn't lift his son because he was so \underline{heavy}. (Answer: son)\\

\hline

2 a) & The older students were bullying the younger ones, so we \underline{punished} them. (Answer: the older students)\\
2 b) & The older students were bullying the younger ones, so we \underline{rescued} them. (Answer: the younger ones)\\

\hline



3 a) & Sam tried to paint a picture of shepherds with sheep, but they ended up looking more like \underline{golfers}. (Answer: shepherds)\\
3 b) & Sam tried to paint a picture of shepherds with sheep, but they ended up looking more like \underline{dogs}. \newline (Answer: sheep)\\

\hline
\end{tabular}
\caption{Examples of Winograd instances. 
}
\label{tab:examples}
\end{center}
\vskip -.1in
\end{table*}

WSC problems are simple for people to solve (human participants in one study performed at 92\% accuracy \cite{bender2015establishing})
but difficult for automatic systems. This is because common sense reasoning encompasses many types of reasoning (causal, spatio-temporal, etc.) and requires a wide breadth of knowledge. 

COPA is a related task that tests a system's ability to recognize causality \cite{roemmele2011choice}. Each  instance comprises a premise and two candidate causes or effects, where the correct choice  is  the candidate that is more \textit{plausible}.



Previous approaches to common sense reasoning, for instance based on logical formalisms \cite{bailey2015winograd} or deep neural models \cite{liu2016probabilistic}, have solved only restricted subsets of the WSC with high precision. They have been tailored for manually selected subsets that demand a specific type of reasoning~\cite{sharma2015towards,liu2016probabilistic}. Others have developed systems for relaxed common sense datasets with looser constraints \cite{rahman2012resolving,peng2015solving,kruengkrai2014example}. In parallel, more general work on common sense reasoning aims to develop a repository of common knowledge using semi-automatic methods (e.g., Cyc \cite{lenat1995cyc} and ConceptNet \cite{liu2004conceptnet}). However, such knowledge bases are necessarily incomplete.

In this work, we propose a general method to resolve common sense problems like WSC and COPA. Contrary to previous work, we aim to solve \emph{all} problem instances rather than a restricted subset. 
Our method is based on on-the-fly \emph{knowledge hunting} and operates in four stages. First, it parses an input problem into a representation schema. Next it generates search queries from the populated schema. It sends these to a search engine, and the next stage parses and filters the results. Finally, it classifies and weighs the results as evidence for respective candidate resolutions.

Our approach arises from the hypothesis that there is too much common sense to encode it all statically; e.g., within a knowledge base or a neural model (using existing techniques).
Even modern NLP corpora composed of billions of words are unlikely to offer good coverage of common sense, or if they do, instances of specific knowledge are likely to be ``long-tailed'' and difficult for statistical systems to model effectively.
Information retrieval (IR) techniques can sidestep these issues by returning targeted results and by using the \emph{entire} indexed internet as a knowledge source.
Scenarios that appear in natural text can offer implicit or explicit evidence for the plausibility of related scenarios in common sense problems.
To solve (\ref{ex-1a}), the following search result contains the relevant knowledge without the original ambiguity:
\begin{exe}
  \ex I got really upset with her and I started to yell at her because... 
\end{exe}
Here, the same entity, \textit{I}, is the subject of both \textit{upset} and \textit{yell at}, which is strong evidence for resolving the original statement. This information can be extracted from a syntactic parse of the retrieved passage with standard NLP tools.

As we will demonstrate experimentally, our knowledge hunting approach achieves an F1 score of 0.51 on the WSC, improving significantly over the previous state-of-the-art (0.3 F1).
When tested on the similar COPA task, a simplified knowledge hunting system performs competitively with the previous best.
To our knowledge, this is the first method that tackles multiple common sense tasks with strong performance on each.
Thus, knowledge-hunting embodies some of the general capabilities that we desire of automatic systems for common sense reasoning.\footnote{Code to reproduce these results are available at https://github.com/aemami1/Wino-Knowledge-Hunter}

\section{Related Work} \label{relwork}

There is increasing interest in using IR approaches to address difficult coreference problems. For example, a recent system \cite{rahman2012resolving} uses web query information to retrieve evidence for the coreference decision in a Winograd-like corpus. 
Other systems~\cite{kobdani2011bootstrapping,ratinov2012learning,bansal2012coreference,zheng2013dynamic,peng2015solving,sharma2015towards} rely on similar techniques, i.e., using search-query counts or co-occurrence statistics and word alignment methods to relate antecedents with pronouns.

Most recent approaches have tackled the Winograd problem by simplifying it in one of two ways. First, systems have been developed exclusively for Rahman and Ng's expanded Winograd-like corpus. These include \citet{rahman2012resolving}'s system itself, achieving 73\% accuracy, and \citet{peng2015solving}'s system (76\%). \citet{kruengkrai2014example} use sentence alignment of web query snippets to achieve 70\% accuracy on a subset of the expanded corpus.
Many instances in this corpus can be resolved using associations between candidate antecedents and the query predicate. For example, ``Lions eat zebras because they are predators.'' Many of the above systems simply query ``Lions are predators'' versus ``zebras are predators'' to make a resolution decision.
This exploitation is often the top contributor to such systems' overall accuracy \cite{rahman2012resolving}, but fails to apply in the majority (if not all) of the original Winograd instances.\footnote{This is why we do not evaluate our method directly on the expanded corpus.}
Our work alleviates this issue by generating search queries that are based exclusively on the predicates of the Winograd instance, not the antecedents, and by considering the strength of the evidence.

Other systems do tackle the original, more difficult Winograd instances, but only a small, author-selected subset. The selection is based often on knowledge-type constraints.
\citet{sharma2015towards}'s knowledge-hunting module focused on a subset of 71 instances that exhibit \emph{causal} relationships; \citet{liu2016probabilistic}'s neural association model focused on a similar causal subset of 70 instances, for which events were extracted manually; and finally, a recent system by \citet{huang2017commonsense} focused on 49 instances.
While these approaches demonstrate that difficult coreference problems can be resolved when they adhere to certain knowledge or structural constraints, they may fail to generalize to other settings. This factor often goes unnoticed when systems are compared only in terms of precision; accordingly, we use an F1-driven comparison that does not enable precision boosting at the cost of recall.

Concurrently with our work,
\citet{trinh2018simple} introduced a system composed of 14 ensembled language models, pre-trained in an unsupervised manner, that achieves up to 63.7\% accuracy on the Winograd Schema Challenge. Compared to our approach, their method requires training multiple language models with vast amounts of data, which is much more expensive.


\paragraph{Other Common-sense Tasks:}
There are various other Turing-test alternatives that directly or indirectly assess common-sense reasoning. These include Pronoun Disambiguation Problems (more generalized, Winograd-like passages without the twist of a special word or twin) \cite{morgenstern2016planning}, the Narrative cloze task \cite{taylor1953cloze}, or its more difficult counterpart, the NarrativeQA Reading Comprehension Challenge \cite{kovcisky2017narrativeqa}.

The COPA task was proposed by \citet{roemmele2011choice}, who also measured the performance of several systems.
The most successful used  Pointwise Mutual Information (PMI) statistics~\cite{church1990word} between  words  in  the  premise and each alternative obtained from a large text corpus (as an implicit way to estimate causal association).
More recent work showed that applying the same PMI-based technique on a corpus of stories yields better results \cite{gordon2011commonsense}.
The current state-of-the-art approaches leverage co-occurrence statistics extracted using causal cues \cite{luo2016commonsense,sasaki2017handling}.

\paragraph{Extended Work:}
Previously, \citet{emami2018generalized} proposed a similar knowledge hunting framework to tackle the Winograd Schema Challenge. This work modifies and extends their approach.
Our modifications include a query-filtering step and various other tweaks that improve results by 0.05 F1 for our best model. In addition, we added further experiments and an ablation study that explores the performance of different model components. Finally, we adapted our method to a new dataset, COPA, on which we achieve respectable results. Accordingly, we change the general takeaway of the previous work from a method with strong performance on a single dataset to one that generalizes and performs well on various tasks.



\section{Knowledge Hunting Framework}
Our framework takes as input a problem instance and processes it through four stages to make a final coreference decision. First, it fits the instance to a semantic representation schema. Second, it generates a set of queries that capture the predicates in the instance's clauses and sends these to a search engine, which retrieves text snippets that closely match the schema. The returned snippets are then parsed and filtered. Finally, the snippets are resolved to their respective antecedents and the results are mapped to a best guess for the original instance's resolution.
We detail these stages below, grounding our description in Winograd instances.
\vskip -.5in
\subsection{Semantic Representation Schema}

The first step is to perform a partial parse of each instance into a shallow semantic representation; that is, a general skeleton of each of the important semantic components in the order that they appear. This is performed using rules related to the syntactic parse of the sentence determined by Stanford CoreNLP \cite{manning2014stanford}.

In general, Winograd instances can be separated into a \textit{context} clause, which introduces the two competing antecedents, and a \textit{query} clause, which contains the target pronoun to be resolved.
We use the following notation to define the components in our representation schema:
\begin{align*}
& E_1, E_2   &  \text{the candidate antecedents} \\
& Pred_C     &  \text{the context predicate} \\
& +          &  \text{discourse connective} \\
& P          &  \text{the target pronoun} \\
& Pred_Q     &  \text{the query predicate} \\
\end{align*}
$E_1$ and $E_2$ are noun phrases in the sentence. In the WSC, these two are specified without ambiguity. $Pred_C$ is the context predicate  composed of the verb phrase that relates both antecedents to some event. The context contains $E_1$, $E_2$, and the context predicate $Pred_C$. The context and the query clauses are often connected by a discourse connective $+$. The query contains the target pronoun, $P$, which is also specified unambiguously. In addition, preceding or succeeding $P$ is the query predicate, $Pred_Q$, a verb phrase involving the target pronoun. Table~\ref{tab:representation} shows sentence pairs in terms of each of these components.

\begin{table*}[ht]
\small 
\begin{center}
\begin{tabular}{L{0.5cm}L{1.5cm}L{1.5cm}L{1.5cm}L{2cm}L{0.8cm}L{3cm}} 
\toprule

Pair & $Pred_C$ & $E_1$ & $E_2$ & $Pred_Q$ & $P$ & Alternating Word (POS)  \\

\midrule

1 & couldn’t lift & the man & his son & was so heavy & he & weak/heavy (adjective) \\

\hline

2 & were bullying & the older students &the younger ones & punished & them & punished/rescued (verb) \\

\hline

3 & tried to paint & shepherds & sheep & ended up looking more like & they & golfers/dogs (noun) \\

\bottomrule
\end{tabular}
\caption{Winograd sentence pairs from Table~\ref{tab:examples}, parsed into the representation schema that we define.}
\label{tab:representation}
\end{center}
\end{table*}

\begin{table*}[h]
\small 
\centering
\begin{tabular*}{\linewidth}{p{4.0cm}p{5.5cm}p{5.0cm}}
\toprule
\multicolumn{3}{l}{\textbf{Sentence}: The trophy doesn't fit into the brown suitcase because it is too large.} \\ \bottomrule \toprule

\textbf{Query Generation Method} & $C$ & $Q$ \\ \midrule

Automatic                & \{``doesn't fit into'', ``brown'', ``fit'' \}                            & \{``large'', ``is too large''\} \\ 
Automatic, with synonyms & \{``doesn't fit into'', ``brown'', ``accommodate'', ``fit'', ``suit'' \} & \{``large'', ``big'', ``is too large'' \} \\
Manual                   & \{``doesn't fit into'', ``fit into'',``doesn't fit'' \}                  & \{``is too large'', ``too large'', ``large'' \}   \\
\bottomrule
\end{tabular*}
\caption{Query generation techniques on an example Winograd sentences}
\label{tab:query-examples}
\vskip -.1in
\end{table*}

\subsection{Query Generation}
Based on the parse, the system generates queries to send to a search engine.
The goal is to retrieve text snippets that resemble the original instance.
Queries are of the form:

$+Term_C$ $+Term_Q$ $-$``Winograd''$-E_1$

\noindent We assume that the search queries are composed of two components, $Term_C$ and $Term_Q$, which are strings that represent the events occurring in the first (context) and second (query) clause of the sentence, respectively. By excluding search results that may contain \textit{Winograd} or $E_1$, we ensure that we do not cheat by retrieving some rewording of the original Winograd instance. 

The next task is to construct two query sets, $C$ and $Q$, whose elements are possible entries for $Term_C$ and $Term_Q$, respectively.
We identify the root verbs in the context and query clauses, along with any modifying adjective, using the dependency parse of the sentence determined by Stanford CoreNLP \cite{manning2014stanford}.
We add the root verbs and adjectives into the sets $C$ and $Q$ along with their broader verb phrases (again identified directly using the dependency tree).

\paragraph{Augmenting the query set with WordNet}
We use WordNet \cite{kilgarriff2000wordnet} to construct an augmented query set that contains synonyms for the verbs and adjectives involved in a representation. In particular, we include the synonyms listed for the top synset of the same part of speech as the extracted verb or adjective. 

\paragraph{Query filtering}
Automated query generation sometimes yields terms that are irrelevant to the disambiguation task. This can add noise to the results. To address this, we implement a semantic similarity algorithm that filters root verbs and modifying adjectives from the query sets according to their relevance to other terms. We estimate relative relevance using Wu-Palmer~\cite{wupalmer} similarity scores from WordNet and filter as follows.
For each passage, the semantic filter (i) computes similarity scores for every possible combination of  $\left\{Term_C, Term_Q\right\}$ (if both $Term_C$ and $Term_Q$ are single words); (ii) determines the maximum similarity score $s$; and (iii) discards any term whose highest similarity score from step (i) is less than $\alpha s$, where $0 < \alpha < 1$. We tune $\alpha$, a hyperparameter,
on Rahman and  Ng’s expanded corpus \cite{rahman2012resolving}.

We hypothesize that terms in the query and context clauses more pertinent to the task have higher mutual similarity scores than irrelevant terms.
To illustrate this, consider the query sets generated for Example 2a, Table 1: \{``bullying'', ``younger'', ``older''\}  and \{``punished''\}. Applying the semantic filter yields the new sets \{``bullying''\} and \{``punished''\}, where the irrelevant terms \textit{younger} and \textit{older} have been removed.

\paragraph{Manual query construction} To understand the impact of the query generation step, we also manually produced representations for all Winograd instances. We limited the size of these sets to five to prevent a blowing-up of search space during knowledge extraction. In Table~\ref{tab:query-examples}, we show examples of generated queries for $C$ and $Q$ using the various techniques.

\subsection{Parsing the Search Results}
From the search results, we obtain a set of text snippets that we filter for similarity to the original problem instance. First, $Term_C$ and $Term_Q$ are restricted to occur in the same snippet, but are allowed to occur in any order. We filter the passed sentences further to ensure that they contain at least two entities that corefer.
These may be structured as follows:
\\ \\
\begin{tabular}{lcl}
   ~ $E_1' ~ Pred_C' ~ E_2'$ &$+$& $E_3' ~ Pred_Q'$ \\
   ~ $E_1' ~ Pred_C' ~ E_2'$ &$+$& $Pred_Q' ~ E_3'$\\
   ~ $E_1' ~ Pred_C'$ &$+$& $E_3'~Pred_Q'$\\
   ~ $E_1' ~ Pred_C'$ &$+$& $Pred_Q' ~ E_3'$\\
\end{tabular}\\ \\
We call these \emph{evidence sentences}. They exhibit a structure similar to the corresponding Winograd instance, but with different entities and event order. $Pred_C'$ and $Pred_Q'$ (resulting from the queries $Term_C$ and $Term_Q$, resp.) should be similar if not identical to $Pred_C$ and $Pred_Q$ from the Winograd sentence. However, $E_1'$, $E_2'$, and $E_3'$ may not have the same semantic type, potentially simplifying their coreference resolution. A sentence for which $E_3'$ refers to $E_1'$ is subsequently labelled \emph{evidence-agent}, and one for which $E_3'$ refers to $E_2'$, \emph{evidence-patient}. The exception to this rule is when an event occurs in the passive voice (e.g., \textit{was called}), which reverses the conventional order of the agent and patient. Another exception is in the case of \textit{causative alternation}, where a verb can be used both transitively and intransitively. The latter case can also reverse the conventional order of the agent and patient (e.g., \textit{he opened the door} versus \textit{the door opened}).
\\ \indent As an example of coreference simplification, a valid evidence sentence is: \textit{He tried to call \textbf{her} but \textbf{she} wasn't available.} Here, the sentence can be resolved on the basis of the gender of the antecedents;  $E_3'$ (the pronoun \textit{she}) refers to the patient, $E_2'$. Accordingly, the sentence is considered an evidence-patient.

\subsection{Antecedent Selection}

We collect and reason about the set of retrieved sentences using a selection process that (i) resolves $E_3'$ to either $E_1'$ or $E_2'$ using CoreNLP's coreference resolution module (rendering them evidence-agent or evidence-patient); and (ii) uses both the count and individual features of the evidence sentences to resolve the original Winograd instance.
For example, the more similar evidence-\textbf{agents} there are for the sentence \textit{Paul tried to call George on the phone, but he wasn't successful}, the more likely it is that the process would guess \textit{Paul}, the \textbf{agent}, to be the correct referent of the target pronoun. \\
\indent To map each sentence to either an evidence-agent or evidence-patient, we developed a rule-based algorithm that uses the syntactic parse of an input sentence.
This algorithm outputs an evidence label along with a list of features. The features indicate: which two entities co-refer according to Stanford CoreNLP's resolver, and to which category of $E_1'$, $E_2'$, or $E_3'$ each belong; the token length of the sentence's search terms, $Term_C$ and $Term_Q$; the order of the sentence's search terms; whether the sentence is in active or passive voice; and whether or not the verb is causative alternating.
Some of these features are straightforward to extract (like token length and order, and coreferring entities given by CoreNLP), while others require various heuristics.
To map each coreferring entity in the snippet to $E_1'$, $E_2'$, or $E_3'$ (corresponding loosely to context subject, context object, and query entity, respectively), we consider their position relative to the predicates in the original Winograd instance. That is, $E_1'$ precedes $Term_C$, $E_2'$ succeeds $Term_C$, and $E_3'$ may precede or succeed $Term_Q$ depending on the Winograd instance. To determine the voice, we use a list of auxiliary verbs and verb phrases (e.g., \textit{was}, \textit{had been}, \textit{is}, \textit{are being}) that switch the voice from active to passive (e.g., ``they are being bullied" vs ``they bullied") whenever one of these precedes $Term_C$ or $Term_Q$ (if they are verbs). Similarly, to identify causative alternation, we use a list of causative alternating verbs (e.g., \textit{break}, \textit{open}, \textit{shut}) to identify the phenomenon whenever $Term_C$ or $Term_Q$ is used intransitively.

These features determine the evidence label, evidence-agent (EA) or evidence-patient (EP), according to the following rules:
\[
    \textit{L}(e)= 
\begin{cases}
    \text{EA},& \text{if } \text{$E_3'$ refers to $E_1'$, active (1)}\\
    \text{EA},& \text{if } \text{$E_3'$ refers to $E_2'$, passive (2)}\\
    \text{EP},& \text{if } \text{$E_3'$ refers to $E_2'$, active (3)}\\
    \text{EP},& \text{if } \text{$E_3'$ refers to $E_1'$, passive (4)}\\
    \text{EP},& \text{if } \text{$E_3'$ refers to $E_1'$, causative (5)}\\
\end{cases}
\]
Cases (2), (4), and (5) account for the passive and causative constructions, which alter the mapping from syntactic role to semantic role.


In addition to determining the evidence label, the features are  used in a heuristic that generates scores (called \textit{strengths}) for each evidence sentence: 
\[
\textit{Str}(e)=\textit{LenScore}(e)+\textit{OrderScore}(e)
\]
\[
    \textit{LenScore}(e)= 
\begin{cases}
    2,& \text{if } len(Term_Q) > 1\\
    2,& \text{if } len(Term_C) > 1\\
    1,              & \text{otherwise}
\end{cases}
\]

\[
    \textit{OrderScore}(e)= 
\begin{cases}
    2,& \text{if } Term_C \prec Term_Q\\
    1,& \text{if } Term_Q \prec Term_C
\end{cases}
\]

As an example of scoring for an actual snippet, let us consider ``She \textbf{tried to call} for him and then search for him herself, but \textbf{wasn’t successful},'' returned for $Term_C$=\textit{tried to call}, and $Term_Q$=\textit{wasn't successful}.

Here, both $Term_Q$ and $Term_C$ are multi-word search terms, and $Term_C$ precedes $Term_Q$ as in the original Winograd sentence. Its overall evidence strength is 4, the highest possible score. On the other hand, for the retrieved snippet ``Has your husband \textbf{tried} Sudafed and was it \textbf{successful}?'' for $Term_Q$=\textit{tried}, and $Term_C$=\textit{successful}, the evidence strength would be 3.
We designed the scoring system to capture the structural similarity of a snippet to its corresponding Winograd instance. We observed that a greater quantity of snippets can be retrieved for less specific search terms, but with increasing noise; we sought to account for this with the features described above.
Note also that our use of the word \textit{features} is intentional. While the weights assigned for the length and order scores could be optimized, as parameters, we consider it inappropriate to do so on the WSC since it is widely used as a \emph{test set}. We set these weights according to our best guess and validated our choices through experiments on the set of Winograd-like sentences provided in \citet{rahman2012resolving}.

We run the above four processes on all snippets retrieved for the input Winograd instance.
The sum of strengths for the evidence-agents is finally compared to that of the evidence-patients to make the resolution decision.

\section{Experiments and Results}

We tested several versions of our framework on the original 273 Winograd sentences (135 pairs and one triple).
These vary in the method of query generation: automatic \textit{vs.} automatic with synonyms \textit{vs.} manual. We compared these systems with previous work on the basis of Precision (P), Recall (R), and F1. 

We used Stanford CoreNLP's coreference resolver \cite{raghunathan2010multi} during query generation to identify the predicates from the syntactic parse, as well as during antecedent selection to retrieve the coreference chain of a candidate evidence sentence. Python's Selenium package was used for web-scraping and Bing-USA and Google (top two pages per result) were the search engines. The search results comprise a list of document snippets that contain the queries (for example, ``yelled at'' and ``upset''). We extract the sentence/s within each snippet that contain the query terms, with the added restriction that the terms should be within 70 characters of each other to encourage relevance.


\begin{table}[ht]
\begin{center}
\begin{tabular}{l|l|l|l|l|}
\cline{2-5}
                                        & \# Correct & P    & R    & F1   \\ \hline
\multicolumn{1}{|l|}{AGQ}               & 77        & 0.56 & 0.28 & 0.38 \\ \hline
\multicolumn{1}{|l|}{AGQ+F}               & 80        & 0.63 & 0.29 & 0.40 \\ \hline
\multicolumn{1}{|l|}{AGQS} & 114        & 0.57 & 0.42 & 0.48 \\ \hline
\multicolumn{1}{|l|}{AGQS+F} & 119        & 0.60 & \textbf{0.44} & \textbf{0.51} \\ \hline
\multicolumn{1}{|l|}{S2015}             & 49         & \textbf{0.92} & 0.18 & 0.30 \\ \hline 
\multicolumn{5}{l}{\textbf{Systems with manual information:}} \\ \hline
\multicolumn{1}{|l|}{L2017}             & 43         & 0.61 & 0.15 & 0.25 \\ \hline 
\multicolumn{1}{|l|}{MGQ}               & 118        & 0.60 & 0.43 & 0.50 \\ \hline
\end{tabular}
\caption{Coverage and performance on the original Winograd Schema Challenge (273 sentences).}
\label{tab:performance}
\end{center}

\end{table}

\begin{table*}[t]
\small 
\begin{center}
\begin{tabular}{cL{12.5cm}} 
\hline

WSC Instance: & The man couldn't lift his son because he was so \underline{weak}. Answer: the man (\textbf{Agent}) \\
\hline 

Evidence and labels: & ``However I \textbf{was so weak} that I \textbf{couldn't lift}"$~\rightarrow~$\textbf{EA}\\
(query terms in bold) & ``She \textbf{was so weak} she couldn't \textbf{lift}"$~\rightarrow~$\textbf{EA}\\
& ``I could not stand
without falling immediately and I \textbf{was so weak} that I \textbf{couldn't lift}"$~\rightarrow~$\textbf{EA}\\
& ``It hurts to \textbf{lift} my leg and its kind of \textbf{weak}"$~\rightarrow~$\textbf{EP}
\\

\hline
Stats and resolution: &  Agent evidence strength: 97 \\
& Patient evidence strength: 72 \\
& Number of scraped sentences: 109 \\
& Resolution: \textbf{Agent} \\

\hline

\end{tabular}
\caption{Example Resolution for a WSC problem. 
}
\label{tab:ResolutionExamples}
\end{center}
\vskip -.2in
\end{table*}

Table~\ref{tab:performance} shows the precision, recall, and F1 of our framework's variants: automatically generated queries (AGQ), automatically generated queries with synonyms (AGQS), and manually generated queries (MGQ). 
We test the automatic systems with (+F) and without the semantic similarity filter.
We compare these to the systems of \citet{sharma2015towards} (S2015) and \citet{liu2017combing} (L2017). The system developed by \citet{liu2017combing} uses elements extracted manually from the problem instances, so is most closely comparable to our MGQ method. Our best automated framework, AGQS+F, outperforms S2015 by 0.21 F1, achieving much higher recall (0.44 vs 0.18). Our results show that the framework with manually generated queries (MGQ) performs better than its automatic counterpart, AGQ, with an F1 of 0.50. AGQS+F slightly outperforms MGQ despite being fully automatic. 

The power of our approach lies in its generality, i.e., its improved coverage of the problem set. It produces an answer for over 70\% of instances. This surpasses previous methods, which only admit specific instance types, by nearly 50\%.

The random baseline on this binary task achieves a P/R/F1 of 0.5. We can artificially raise the F1 performance of all systems above 0.5 by randomly guessing an answer in cases where the system makes no decision. For AGQS+F, for example, if we take a random decision on the cases (74) with no retrieved evidence, we get an accuracy of 57.1\%.
However, we think it is important that systems are compared transparently based on which instances they admit and when they are capable of making a prediction.

\section{Error Analysis}
To get a sense of the performance of our heuristics in classifying evidence sentences in the antecedent selection step, we manually labelled sentences retrieved by the AGQS system for 40 Winograd instances. 
The categories are evidence-agent, evidence-patient, or neither (insufficient evidence). This amounts to a total of 876 evidence sentences. We compared these labels to those assigned by our system. In total, 703 of the 876 evidence sentences were labelled correctly (81\%). Of the 173 incorrect cases, 110 were marked as insufficient evidence. Our system is forced to label these as agent or patient.

Evidence sentences were insufficient for a variety of reasons. Most frequently, they were structurally incomplete or grammatically incorrect, despite passing as valid through CoreNLP and our initial coreference heuristics.
In general, our coreference heuristics filter strongly: over all Winograd instances, they filter a total of 50,110 retrieved sentences down to only 3,097 (0.0617 acceptance rate). As for the 63 cases of sufficient evidence sentences that were labelled incorrectly, the issue was either errors in the coreference information from the CoreNLP pipeline or errors in our heuristics for reasoning about the coreference information. We show examples of these various sources of error in supplementary Table S1. At any rate, the corrected labels (with the 110 insufficient evidence removed and the 63 cases corrected) did not result in a shift in any of the 40 coreference decisions.

In Table~\ref{tab:ResolutionExamples}, we show a sample resolution that our system makes on a problem instance,\footnote{We provide more examples in a supplementary file.} including some evidence that was retrieved and labelled automatically and the evidence strengths that led to the resolution. These examples reveal that, indeed, general knowledge of what is plausible appears in natural text. Our system successfully leverages this knowledge for common sense reasoning.

We also include an example evidence snippet that yields a ``misleading'' label.
Generally, sources of misleading snippets include incomplete or imprecise query generation (e.g. in Table~\ref{tab:ResolutionExamples}, querying only  ``lift'' instead of ``couldn't lift''), errors in the automatic parsing of sentences (e.g., in supplementary Table S1.1.b, ``lift'' is incorrectly labelled as a verb via the parse tree, despite being a noun),
and insufficient filtering of noisy sentences that are not relevant to the problem instance or are incomplete (e.g. in supplementary Table S1.2.b, the sentence is incomplete and indicates a misleading resolution).

\section{Generalization to COPA}
To investigate the generality of our knowledge-hunting approach, we adapted it to the Choice of Plausible Alternatives (COPA).
We evaluated our basic automatic models that did not use the semantic similarity filter for this check.

COPA has a slightly different form that necessitates some modifications. As an example,
\begin{exe}
\ex The climbers reached the peak of the mountain. What happened as a \underline{result}?\begin{xlist} 

    \ex \label{ex-3a} They encountered an avalanche.
    \ex \label{ex-3b} \textbf{They congratulated each other.}
    \end{xlist}
\end{exe}
During query generation, as before, the set $C$ contains terms extracted from the context sentence. Instead of a single set $Q$ as in the WSC, we generate two query sets $Q_1$ and $Q_2$, that contain terms extracted for the first and second candidate sentences.
Because entities in the candidate sentences can contribute to the answer (unlike in the WSC), we modified the query generation rules to extract more than just predicates.
Specifically, the extraction procedure uses the syntactic parse tree of the phrase to back-off from extracting the clause containing the subject and verb phrase, to only the verb phrase, to only the verbs or adjectives that are rooted in the verb phrase.
For the running example, our system generates these three sets: $C$=\{``The climbers reached the peak'', ``reached the peak'', ``reached''\}, $Q_1$=\{``They encountered an avalanche'', ``encountered an avalanche'', ``encountered''\}, and $Q_2$=\{``They congratulated each other'', ``congratulated each other'', ``congratulated''\}.

We query the web for sentences that contain terms in $(C,Q_1)$ and $(C,Q_2)$, with one added restriction: for problem instances in which the relation is \textit{cause}, the system only extracts sentences in which $Term_C$ precedes $Term_{Q_1}$ or $Term_{Q_2}$; when the relation is \textit{result} (as in our running example), $Term_C$ succeeds $Term_{Q_1}$ or $Term_{Q_2}$.
As for the WSC, the final decision is determined from the evidence snippets according to their strengths.



\begin{table}[ht]
\begin{center}
\begin{tabular}{l|p{1.5cm}|p{1.5cm}|}
\cline{2-3}
                                        &  Dev  & Test    \\ \hline
\multicolumn{1}{|l|}{\citet{goodwin2012utdhlt}}             & --         & 63.4   \\ \hline
\multicolumn{1}{|l|}{AGQS}               & 64.0        &  65.1    \\ \hline
\multicolumn{1}{|l|}{\citet{gordon2011commonsense}} & 62.8        & 65.4   \\ \hline
\multicolumn{1}{|l|}{AGQ}               & 65.8        &  66.2\footnotemark    \\ \hline

\multicolumn{1}{|l|}{\citet{luo2016commonsense}}             & --         & 70.2   \\ \hline 

\multicolumn{1}{|l|}{\citet{sasaki2017handling}}               & --        & \textbf{71.2}  \\ \hline

\end{tabular}
\caption{Model accuracy (\%) on COPA.}
\label{tab:COPA}
\end{center}
\vskip -.2in
\end{table}
\footnotetext{This precision can be inflated to 67.2 by randomly guessing on the 10 examples for which there were no search results.}

We tuned the system's evidence-scoring heuristics on COPA's 500 validation instances.  In Table~\ref{tab:COPA}, we compare our system's performance on the 500 test instances to previous work on the basis of precision (which in the full-coverage case equates to accuracy).
Our simpler AGQ method achieves 66.2\% accuracy, which is respectable, although not state-of-the-art. As indicated by the lower performance of AGQS, synonyms from WordNet did not improve performance on COPA. Without the semantic-similarity filtering, synonyms may add noise to the retrieved results. It has also been shown that multi-word expressions are prevalent and important for COPA~\cite{sasaki2017handling}, which we have not specifically attempted to handle with our method. We believe that this is a promising direction of improvement for our approach in future work.




\section{Conclusion}
We developed a knowledge-hunting framework to tackle the Winograd Schema Challenge, a task that requires common-sense knowledge and reasoning.
Our system involves a semantic representation schema and an antecedent selection process that acts on web-search results. We evaluated the performance of our framework on the original set of WSC instances, achieving F1-performance that significantly exceeded the previous state-of-the-art.
A simple port of our approach to COPA suggests that it has the potential to generalize.
In the future we will study how this common-sense reasoning technique can contribute to solving ``edge cases'' and difficult examples in more general coreference tasks.


\section*{Acknowledgements}
This work was supported by the Natural Sciences and Engineering Research Council of Canada. 

\bibliography{emnlp2018}
\bibliographystyle{acl_natbib_nourl}

\end{document}